\documentclass[sigconf]{acmart}
\AtBeginDocument{%
  }

\copyrightyear{2024} 
\acmYear{2024} 
\setcopyright{acmlicensed}\acmConference[KDD '24]{Proceedings of the 30th
ACM SIGKDD Conference on Knowledge Discovery and Data Mining}{August
25--29, 2024}{Barcelona, Spain}
\acmBooktitle{Proceedings of the 30th ACM SIGKDD Conference on Knowledge
Discovery and Data Mining (KDD '24), August 25--29, 2024, Barcelona, Spain}
\acmDOI{10.1145/3637528.3671910}
\acmISBN{979-8-4007-0490-1/24/08}

\usepackage{graphicx}
\usepackage{xcolor}
\usepackage{multirow}
\usepackage{tabularx}
\usepackage{algorithm}
\usepackage{subcaption}
\usepackage{algorithmic}
\usepackage{amsmath}
\usepackage{caption}
\usepackage{multirow}
\usepackage{amsthm}
\usepackage{enumitem}

\newcolumntype{C}{>{\centering\arraybackslash}X} 

\newtheorem{problem}{Problem}

\begin{document}

\title{
Rethinking Graph Backdoor Attacks: A Distribution-Preserving Perspective
}

\author{Zhiwei Zhang}
\affiliation{%
  \institution{The Pennsylvania State University}
  \city{State College}
  \country{USA}
  }
\email{zbz5349@psu.edu}

\author{Minhua Lin}
\affiliation{%
  \institution{The Pennsylvania State University}
  \city{State College}
  \country{USA}
}
\email{mfl5681@psu.edu}

\author{Enyan Dai}
\affiliation{%
  \institution{The Pennsylvania State University}
  \city{State College}
  \country{USA}
}
\email{emd5759@psu.edu}

\author{Suhang Wang}
\affiliation{%
  \institution{The Pennsylvania State University}
  \city{State College}
  \country{USA}
}
\email{szw494@psu.edu}

\begin{abstract}
Graph Neural Networks (GNNs) have shown remarkable performance in various tasks. However, recent works reveal that GNNs are vulnerable to backdoor attacks. Generally, backdoor attack poisons the graph by attaching backdoor triggers and the target class label to a set of nodes in the training graph. A GNN trained on the poisoned graph will then be misled to predict test nodes attached with trigger to the target class. Despite their effectiveness, our empirical analysis shows that triggers generated by existing methods tend to be out-of-distribution (OOD), which significantly differ from the clean data. Hence, these injected triggers can be easily detected and pruned with widely used outlier detection methods in real-world applications. Therefore, in this paper, we study a novel problem of unnoticeable graph backdoor attacks with in-distribution (ID) triggers. To generate ID triggers, we introduce an OOD detector in conjunction with an adversarial learning strategy to generate the attributes of the triggers within distribution. To ensure a high attack success rate with ID triggers, we introduce novel modules designed to enhance trigger memorization by the victim model trained on poisoned graph. Extensive experiments on real-world datasets demonstrate the effectiveness of the proposed method in generating in distribution triggers that can bypass various defense strategies while maintaining a high attack success rate. Our code is available at: \href{https://github.com/zzwjames/DPGBA}{https://github.com/zzwjames/DPGBA}. 
\end{abstract}



\maketitle
\section{Introduction}
Graph-structured data is pervasive in real world, such as social networks \cite{fan2019graph}, molecular structures \cite{Mansimov_2019}, and knowledge graphs \cite{Liu_2022}. With the growing interest in learning from graphs, Graph Neural Networks (GNNs), which have shown great ability in node representation learning on graphs, have become increasingly prominent. Generally, GNNs adopt the message-passing mechanism, which update a node's representation by recursive propagation and aggregation of information from a node's neighbors. The learned node representations preserve both node attributes and local graph structure information, which can benefit a range of downstream tasks, such as node classification \cite{kipf2017semisupervised, veličković2018graph, hamilton2017inductive}, graph classification \cite{xu2019powerful}, and link prediction \cite{zhang2018link}.

Though GNNs have achieved remarkable performance across various applications, recent studies \cite{Dai_2023, xi2021graph, zhang2021backdoor} have shown that they are vulnerable to backdoor attacks. Generally, backdoor attacks generate and attach backdoor triggers to a selected group of nodes, known as target nodes, and assign target nodes a target class. Triggers are typically a node or a subgraph and can either be predefined or generated by a trigger generator. When a GNN model is trained on a dataset poisoned with these triggers, it learns to associate the presence of the trigger with the target class. Consequently, during inference, the backdoored model will misclassify test nodes attached with the trigger to the target class, while maintain high prediction accuracy on clean nodes, i.e., nodes without triggers attached. 
Backdoor attacks on graphs pose a significant threat to the adoption of GNNs in real-world, especially on high-stake scenarios 
such as banking systems and cybersecurity. For example, an adversary could inject backdoor triggers to the training data for fraud detection in transaction networks, and bypass the detection of a model trained on such poisioned graph by disguising illegal behaviors with backdoor triggers. 

Hence, graph backdoor attack is attracting increasing attention and several initial efforts have been taken~\cite{zhang2021backdoor,xi2021graph,Dai_2023}. For example, SBA \cite{zhang2021backdoor} conducts pioneering research on graph backdoor attacks. It adopts randomly generated graphs as triggers. Building on this work, GTA \cite{xi2021graph} adopts a backdoor trigger generator to generate more powerful sample-specific triggers to improve the attack success rate. \citeauthor{Dai_2023}~\cite{Dai_2023} shows that the generated triggers in previous work can be easily broken by pruning edges linking nodes with low cosine feature similarity. To alleviate this issue, they propose UGBA, which adopts an unnoticeable constraint to make the triggers and the target nodes to have large cosine similarity of features. 

Despite their superior attack performance and the initial efforts to make backdoor attack unnoticeable, our preliminary analysis in Sec. \ref{preana} shows that the triggers generated by existing generator-based backdoor attack methods are typically \textit{out-of-distribution} samples compared to the clean data, i.e., 
the feature vectors of the triggers are easily distinguishable from those of clean nodes. 
This is because the victim model, trained on a poisoned dataset, tends to memorize outlier triggers or associate outlier triggers with the target class more easily than in-distribution triggers. Consequently, when the trigger generator is trained without any constraints, it naturally exploits this shortcut to achieve higher attack success rate. 
Though UGBA aims to learn triggers that have large cosine feature similarity with target nodes, it does not take the magnitude of triggers into consideration, resulting in triggers having large features for higher attack success rate. 
This ``out-of-distribution'' property can be leveraged by outlier detection methods to identify and remove those triggers, thus significantly degrading the attack performance. As shown in our preliminary analysis in Sec. \ref{preana}, with an unsupervised outlier detection, we can successfully remove/break the triggers in a poisoned graph, degrading the attack success rate from over $90\%$ to $0\%$ on Pubmed dataset.
As outlier detection is widely deployed in real-world applications such as financial networks \cite{huang2018codetect} and cybersecurity \cite{tuor2017deep}, the out-of-distribution issue undermines the real-world adoption value of these backdoor attack methods. 
For instance, in financial networks, outlier detection methods play a pivotal role in unveiling unusual transaction patterns that may indicate money laundering activities \cite{ahmed2016survey}. 

Developing in-distribution triggers that mimic legitimate patterns within these networks is promising to fool existing outlier detection methods. For instance, in a social network, an ID trigger could replicate the typical behavior patterns of genuine user accounts, making it more difficult for outlier detection methods to distinguish between legitimate activities and those designed to compromise the network's integrity. Hence, the development of an effective graph backdoor attack, using in-distribution (ID) triggers capable of bypassing widely deployed outlier detection methods while maintaining a high attack success rate, holds significant importance. However, there is no existing work on this.

Therefore, in this paper, we study a novel and important problem of developing an effective distribution-preserving graph backdoor attack. In essence, we confront two key challenges: (i) how to generate in-distribution triggers that are resistant to commonly employed outlier detection methods in real-world applications; and (ii) making triggers in-distribution might degrade the attack performance as it breaks the shortcut for the victim model to associate the trigger and the target label. How to achieve a high attack success rate with these ID triggers? In an attempt to address these challenges, we proposed a novel framework \underline{D}istribution \underline{P}reserving \underline{G}raph \underline{B}ackdoor \underline{A}ttack (DPGBA). To generate ID triggers, we introduce an OOD detector and adopt an adversarial learning strategy to constrain the attributes of the generated triggers. In order to enhance the attack success rate utilizing ID triggers, we introduce innovative modules aimed at promoting the memorization of generated triggers by the victim model and encouraging the learned embeddings of the poisoned nodes to resemble those belonging to the target class.
In summary, our main contributions are:
\begin{itemize}[leftmargin=*]
    \item We empirically show that existing backdoor attacks suffer from either low attack success rate or outlier issues that allow outlier detection methods to significantly degrade their performance;
    \item We design a novel graph backdoor attack framework, which can generate in-distribution triggers that can bypass outlier detection and achieve high attack success rate;
    \item Extensive experiments on large-scale dataset demonstrate the effectiveness of our framework in backdooring different GNN models using ID triggers under different defense settings. 

\end{itemize} 
\section{Related Works}
\subsection{Graph Neural Networks}
With the increasing need for learning on graph structured data, Graph Neural Networks (GNNs), which have shown great power in modeling graphs, are developing rapidly in recent years \cite{dai2022towards, Zhao_2021, Zhou_2022}.  
Most GNN variants operate under the message-passing framework, which integrates pattern extraction and interaction modeling across each layer \cite{hamilton2017inductive, zhang2018link, kipf2017semisupervised}. Essentially, GNNs handle messages derived from node representations, propagating these messages through various message-passing mechanisms to enhance node representations. These refined representations are subsequently applied to downstream tasks
With the evolution of GNN technology, numerous advancements have been made to augment their performance and application scope. Innovations in self-supervised learning techniques for GNNs aim to lessen the dependency on annotated data \cite{you2021graph, lin2023certifiably, zhu2020deep, ma2024overcoming, xiao2023graphecl}. Additionally, significant strides have been made in enhancing the fairness \cite{zhu2023fairnessaware, dai2021say}, robustness \cite{dai2022towards} and interpretability of GNN frameworks \cite{zhang2021protgnn}. Furthermore, specialized GNN architectures have been developed to address the unique challenges presented by heterophilic graphs \cite{10027721}, broadening the potential use cases of GNNs in complex networked systems.

\subsection{Backdoor Attacks on Graph}
Backdoor attacks have been widely studied in image domain \cite{li2022backdoor, 8685687, chen2017targeted}.
Initial work directly poison training samples \cite{liu2020reflection, chen2017targeted}. Others have explored the invisibility of triggers \cite{li2021invisible, doan2021backdoor}. 
Besides, the hidden backdoor could also be embedded through transfer learning \cite{kurita2020weight}, modifying model parameters \cite{chen2021proflip}, and adding extra malicious modules \cite{tang2020embarrassingly}.
Recent studies have begun to delve into backdoor attacks on GNNs, focusing on a strategy distinct from the more prevalent poisoning and evasion attacks. Backdoor attacks involve injecting malicious triggers in the training data, which cause the model to make incorrect predictions when these triggers are presented in test samples. This form of attack subtly manipulates the training phase of a model, ensuring it performs as expected under regular conditions but fails in the presence of trigger-embedded inputs. 
Among the pioneering efforts, SBA \cite{zhang2021backdoor} introduced a method for injecting universal triggers into training samples through a subgraph-based approach. However, the attack success rate is poor. 
GTA \cite{xi2021graph} furthered this by developing a technique for generating adaptive triggers, customizing perturbations for individual samples to enhance attack effectiveness. 
In UGBA \cite{Dai_2023}, an algorithm for selecting poisoned nodes is introduced to optimize the utilization of the attack budget. Additionally, an adaptive trigger generator is employed to create triggers that demonstrate a high cosine similarity to the target node. While GTA and UGBA achieve a high attack success rate, the generated triggers tend to be outliers. This is because it is more efficient for a victim model to associate outlier triggers with the target class, leading the unconstrained trigger generator to exploit this shortcut for a higher attack success rate.

The aforementioned graph backdoor methods either have low attack success rate or outlier issues that makes them ineffective in presence of outlier detection.
A detailed review of existing outlier detection on graph is given in Appendix \ref{odrelated}. 
Our proposed method is inherently different from these methods as
(i) we aim to generate unnoticeable in-distribution triggers capable of bypassing the commonly used outlier detection methods in real-world applications.
(ii) we focus on guaranteeing a high attack success rate with these in-distribution triggers.

\section{Preliminaries Analysis}
In this section, we give preliminaries of backdoor attacks on graphs and show out-of-distribution issues of existing backdoor attacks.

\begin{figure*}[ht]
    \centering
    \includegraphics[width=\linewidth]{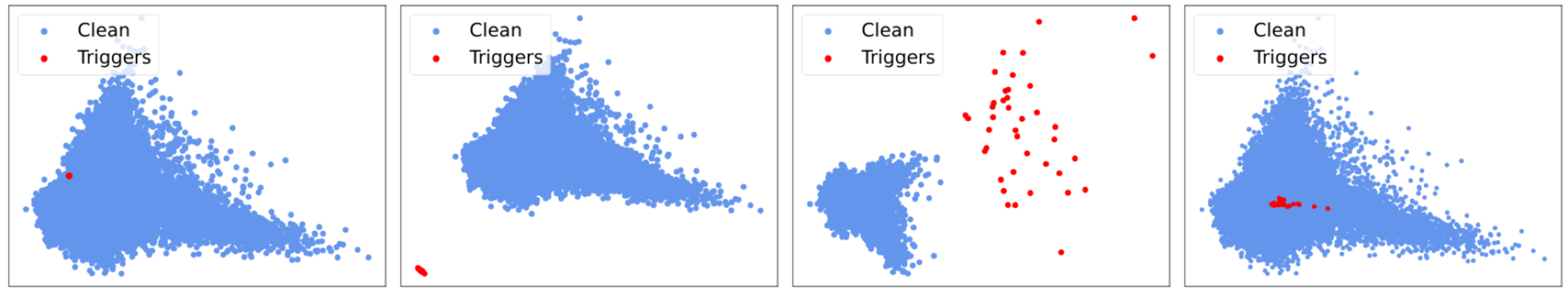}
    \label{fig:main}

    \bigskip 

    \vskip -2em
    \begin{minipage}{0.25\textwidth}
        \centering
        \textbf{(a) SBA}
        \vspace{1mm}
    \end{minipage}%
    \begin{minipage}{0.25\textwidth}
        \centering
        \textbf{(b) GTA}
        \vspace{1mm}
    \end{minipage}%
    \begin{minipage}{0.25\textwidth}
        \centering
        \textbf{(c) UGBA}
        \vspace{1mm}
    \end{minipage}%
    \begin{minipage}{0.25\textwidth}
        \centering
        \textbf{(d) Ours}
        \vspace{1mm}
    \end{minipage}
    \vskip -1.5em
    \caption{PCA visualization of features of clean and generated triggers by different attacks. Red dots are overlapped in (a) as SBA generate the same trigger for all target nodes.} 
    \label{pre}
\end{figure*}

\subsection{Notations}
\label{notation}


We denote an attributed graph as $\mathcal{G} = (\mathcal{V}, \mathcal{E}, \mathbf{X})$, where $\mathcal{V} = \{v_1, \dots, v_N\}$ represents the set of $N$ nodes, $\mathcal{E} \subseteq \mathcal{V} \times \mathcal{V}$ is the set of edges, and $\mathbf{X} = \{\mathbf{x}_1, \dots, \mathbf{x}_N\}$ denotes the set of node attributes, with $\mathbf{x}_i$ being the attribute of node $v_i$. The adjacency matrix of the graph $\mathcal{G}$ is denoted as $\mathbf{A} \in \mathbb{R}^{N \times N}$, where $A_{ij}=1$ if nodes $v_i$ and $v_j$ are connected; otherwise, $A_{ij}=0$.
In this paper, we concentrate on backdoor attack for semi-supervised node classification task within the inductive setting. Specifically, the training graph $\mathcal{G}$ includes a small subset of labeled nodes $\mathcal{V}_L \subseteq \mathcal{V}$ with labels as $ \mathcal{Y}_L=\left \{ y_1,\dots,y_{N_L} \right \} $. The remaining nodes of $\mathcal{G}$ are unlabeled, denoted as $\mathcal{V}_U$. 
We denote $\mathcal{V}_{Tr}=\mathcal{V}_{L}\cup \mathcal{V}_{U}$ as the node set for the training graph.
The test nodes, denoted as $\mathcal{V}_T$, are not part of the training graph $\mathcal{G}$, i.e., $\mathcal{V}_{T} \cap \mathcal{V}_{Tr} = \emptyset$. We aim to add backdoor triggers within budget to the training graph such that a GNN model trained on the backdoored graph will be fooled to give targeted label for test nodes attached with triggers.

\subsection{Preliminaries of Graph Backdoor Attacks}
Next, we elaborate on the attacker's objectives, knowledge, and capabilities, followed by the details of the inductive setting employed for evaluating the attack.

\noindent\textbf{Attacker's Goal}: The attacker aims to add backdoor triggers, i.e., nodes or subgraphs, to a small set of target nodes in the training graph and label them as a target class, such that a GNN model trained on the poisoned graph will memorize the backdoor trigger and be misguided to classify target nodes attached with triggers as the target class. Meanwhile, the attacked GNN model should behave normally for clean nodes without triggers attached.

\noindent\textbf{Attacker’s Knowledge and Capability}: In the context of most poisoning attacks \cite{sun2020adversarial}, attackers have access to the training data of the target model. However, they lack information about the specifics of the target GNN models, including their architecture. Attackers have the capability to attach triggers and labels to nodes within a predefined budget prior to the training of the target models in order to poison the graphs. In the inference phase, attackers retain the ability to attach triggers to the target test nodes.

\noindent\textbf{Evaluation Setting}:
Given $\mathcal{V}_P\subseteq \mathcal{V}_{U}$ as a set of poisoned node,
we attach the generated trigger $g_i = (\mathbf{X}^g_i, \mathbf{A}^g_i)$ to the node $v_i \in \mathcal{V}_P$ and assign $\mathcal{V}_P$ with target class $y_t$ to form the backdoored dataset. The victim model is then trained on this dataset. 
During inference, triggers generated by trigger generator $f_g$ are attached to test nodes $v_i \in \mathcal{V}_T$ to evaluate the attack performance.

\subsection{Outlier Issues of Graph Backdoor Attacks}
\label{preana}
An implicit requirement for backdoor attacks is that the generated triggers should be indistinguishable from clean inputs. This condition is commonly satisfied in the image domain \cite{li2022backdoor} by constraining backdoor triggers to the input, such as using small patch patterns or imperceptible perturbations. However, in the context of backdoor attacks on graphs, where new samples are generated, without specific design to constrain in-distribution trigger generation, the trigger generator may take a shortcut and produce outlier triggers which can be easily memorized by the victim model. Though such triggers have high attack success rate, they are outliers and can be easily removed by simple outlier detection algorithms, making them ineffective in practice. 

To show that the triggers generated by existing graph backdoor attack methods are outliers, we conduct analysis on Pubmed dataset \cite{Sen_Namata_Bilgic_Getoor_Galligher_Eliassi-Rad_2008}. We first adopt existing backdoor attack algorithms to add backdoors to the graph under the semi-supervised setting. We then apply Principal Component Analysis (PCA) to reduce the dimensionality of node attributes for both clean nodes and triggers' nodes, and visualize them in a 2-dimensional space as shown in Fig. \ref{pre}, where 
the blue and the red dots denote the clean node and the generated triggers, respectively. From the figure, it is obvious that the generated triggers of GTA \cite{xi2021graph} and UGBA \cite{Dai_2023} are very different and far from the clean data distribution, showing that the triggers generated by many existing algorithms are outliers.


\begin{table}[t]
\small
\centering
\caption{Results of backdoor defense (Attack Success Rate ($\%$)
| Clean Accuracy ($\%$)) on PubMed dataset.}
\setlength{\extrarowheight}{.065pt}
\vskip -1em
\begin{tabular}{cccccc}
\hline
Defense & Clean & SBA-Samp & SBA-Gen & GTA & UGBA \\ 
\hline
None & 84.9 & 30.4 | 84.7 & 32.0 | 84.6 & 86.6 | 84.9 & 92.3 | 84.9 \\
OD & 84.8 & 29.6 | 84.9 & 31.7 | 84.6 & \ \ 0.0 | 85.0 & \ \ 0.0 | 84.7 \\
\hline 
\end{tabular}
\label{preex}
\end{table}

To show that such triggers are ineffective in practice, i.e., can be easily detected, we employ outlier detection (\textbf{OD}) to defend against existing backdoor attacks. 
Specifically, we adopt DOMINANT \cite{ding2019deep}, a popular unsupervised outlier detection method based on autoencoder for graph-structured data, which utilize the reconstruction error on both graph structural and node attribute for outlier detection. The intuition is that the autoencoder will be better at reconstructing instances that are similar to the majority of the data it was trained on (presumably normal data) and worse at reconstructing outliers \cite{NEURIPS2022_acc1ec4a}. 
Given a backdoored dataset, we train DOMINANT on it and then discard those samples with high reconstruction loss. 
Experiment results on Pubmed \cite{Sen_Namata_Bilgic_Getoor_Galligher_Eliassi-Rad_2008} with $|\mathcal{V}_P|$ set as 40 are presented in Table \ref{preex}. The architectures of the target model is GCN \cite{kipf2017semisupervised} and the size of triggers is limited to contain three nodes. We filter out the top $3\%$ of samples with the highest reconstruction losses. 
More results on other datasets can be found in Table \ref{maintable}. The accuracy of the backdoored GNN on clean test set is also reported in Table \ref{preex} to show how the defense strategy affect the prediction performance. Accuracy on a clean
graph without any attacks is reported as reference. All the results are averaged scores of 5 runs. The details of evaluation protocol is in Sec \ref{protocol}. 
From the table, we can observe:
(i) Both GTA and UGBA exhibit a high attack success rate without a defense method. However, employing a straightforward outlier detection strategy effectively eliminates all of their triggers, degrading the attack success rate to 0; 
(ii) For SBA, which generates triggers based on the mean and standard deviation of the clean input, it achieves a low attack success rate despite its triggers not being classified as outliers. 
Evidently, \textit{existing backdoor attacks methods on graph suffer from either a low attack success rate or outlier issues}. Thus, it is important to design a framework capable of generating ID triggers that can achieve a high attack success rate and bypass outlier detection.


\subsection{Problem Definition}
Our preliminary analysis shows that existing backdoor attacks either have a low attack success rate or encounter outlier issues. 
To address these problems, we propose to develop a novel and effective distribution preserving graph backdoor attack that can generate in-distribution triggers capable of bypassing commonly employed outlier detection techniques, while maintaining a high attack success rate. As we aim to bypass outlier detection techniques, we define the distribution preserving as follows.

\noindent\textbf{In-Distribution Constraint on Triggers.}
\label{idcons}
Let $\mathcal{G}_B=(\mathcal{V}\cup\mathcal{T}_P, \mathcal{E}\cup\mathcal{E}_P, \mathbf{X}\cup\mathbf{X}_P )$ be the backdoored graph, where $\mathcal{T}_P$ represents the set of generated triggers, $\mathcal{E}_P$ denotes the edge set containing edges within the triggers $g_i \in \mathcal{T}_P$ and edges attaching these triggers to nodes $v_i \in \mathcal{V}$, and $\mathbf{X}_P$ represents the node attributes of the generated triggers. Let $f_o$ be an outlier detection model trained on $\mathcal{G}_B$.
Then, our in-distribution constraint on trigger $g_i$ is defined as:
\begin{equation}
    f_o(g_i) < \tau
    \label{thres}
\end{equation}
where $f_o(g_i)$ is the anomaly score of $g_i$ and $\tau$ is a threshold which can be tuned based on datasets. 

Following \cite{Dai_2023}, the clean prediction for a node $v_i$ can be denoted as $f_\theta(v_i)=f_\theta(\mathcal{G}_C^i)$, where $\mathcal{G}_C^i$ is the $K$-hop subgraph centered at $v_i$. 
For a node $v_i$ attached with the trigger $g_i$, the predicted label is denoted as
$f_\theta\left(\tilde{v}_i\right)$, where $\tilde{v}_i= a\left(\mathcal{G}_C^i, g_i\right)$ and 
$a(\cdot)$ being the operation of trigger attachment.
With the above descriptions and notations, the effective distribution preserving graph backdoor attack is formally defined as:
\begin{problem}[Distribution Preserving Graph Backdoor Attack]
    Given a clean attributed graph $\mathcal{G} = (\mathcal{V}, \mathcal{E}, \mathbf{X})$ with a set of nodes $\mathcal{V}_L$ provided with labels $\mathcal{Y}_L$,
    we aim to learn an adaptive trigger generator $f_g(v_i) \rightarrow g_i $. 
    This generator will produce triggers that bypass outlier detection while ensuring that a GNN $f$, trained on the poisoned graph will classify the test node attached with the trigger to the target class $y_t$. 
    This objective is achieved by solving:
    \begin{equation}
        \begin{aligned}
& \min _{\theta_g} \sum_{v_i \in \mathcal{V}_U} l\left(f_{\theta_s^*}\left(\tilde{v}_i\right), y_t\right) \\
& \text { s.t. } \theta_s^*=\underset{\theta_s}{\arg \min } \sum_{v_i \in \mathcal{V}_L} l\left(f_s\left(v_i\right), y_i\right)+\sum_{v_i \in \mathcal{V}_P} l\left(f_s \left(\tilde{v}_i\right), y_t\right), \\
& \quad \forall v_i \in \mathcal{V}_P \cup \mathcal{V}_U, g_i\ \text{meets Eq. \eqref{thres} and }\left|g_i\right|<\Delta_g\\
& \quad\left|\mathcal{V}_P\right| \leq \Delta_P
\end{aligned}
        \label{eq:obj}
    \end{equation}
    where $l(\cdot)$ is the cross entropy loss, $y_t$ is the target class label and $\theta_g$ denotes the parameters of the adaptative trigger generator $f_g$. In the constraints, the node size of trigger $|g_i|$ is limited by $\Delta_g$, and the size of poisoned nodes is limited by $\Delta_p$. The architecture of the target GNN $f$ is unknown. Hence, a surrogate GNN classifier $f_s$ with parameters $\theta_s$ is used. 
\end{problem}


\section{Methodology}
\begin{figure}[t]
  \centering
  \includegraphics[width=\linewidth]{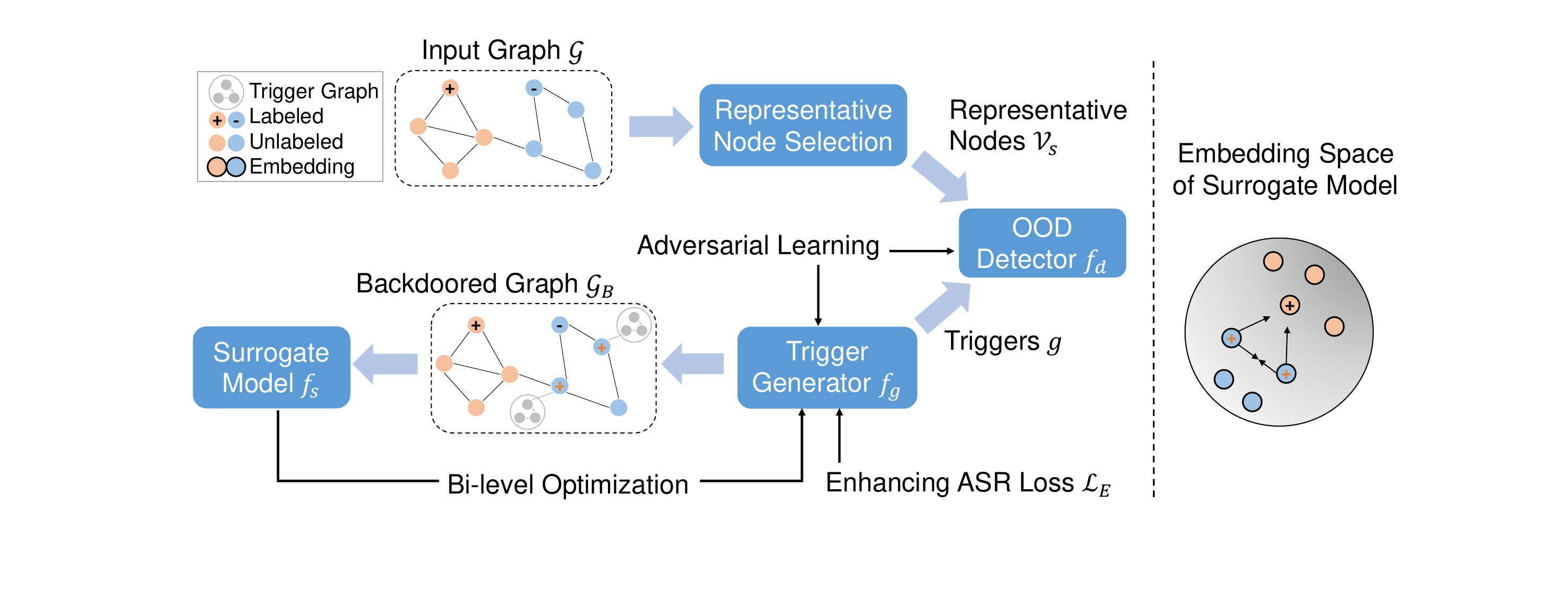}
  \vskip -1em
  \caption{ Framework of DPGBA
  }
  \label{framework}
\end{figure}

In this section, we present the details of the proposed framework, which aims to optimize Eq. \eqref{eq:obj} to conduct effective distribution preserving graph backdoor attacks.
Two challenges remain to be addressed: 
(i) how to generate ID triggers that have the capability to bypass outlier detection defense methods;
(ii) how to learn the trigger generator to obtain triggers that meet ID constraint while maintaining a high attack success rate. 
To address these challenges, a novel framework DPGBA is proposed, which is illustrated in Fig. \ref{framework}. 
DPGBA is composed of an OOD detector $f_d$, a trigger generator $f_s$ and a surrogate node classifier $f_s$.
Specifically, to address the first challenge, an adversarial training strategy involving an OOD detector $f_d$ and a trigger generator $f_g$ is introduced. The OOD detector is trained to differentiate between clean data from a graph $\mathcal{G}$ and triggers generated by $f_g$. In turn, the trigger generator enhances its capability to create triggers that closely mimic the clean data.
To address the second challenge, we propose novel objective functions that promote generated triggers to exert a dominant influence on target nodes. This encourages the victim model to memorize these triggers, leading to a high attack success rate. 
Next, we give the details of each component.

\subsection{In-Distribution Triggers Generation}



In this subsection, we detail the design of the trigger generator $f_g$ and the adversarial learning strategy proposed to ensure the generator produces triggers that are in-distribution.

To make the trigger generator more effective and flexible in generating triggers, instead of using predefined triggers, following \cite{Dai_2023}, we use an MLP as the adaptive trigger generator $f_g$, which generate triggers based on the target node's node attributes. Specifically, given a node $v_i$, $f_g$ generates the node features and graph structure of a trigger to be attached to $v_i$ as:
\begin{equation}
    \mathbf{h}_i^m = \operatorname{MLP}(\mathbf{x}_i), \quad \mathbf{X}_i^g = \mathbf{W}_f \mathbf{h}_i^m, \quad \mathbf{A}_i^g = \mathbf{W}_a \mathbf{h}_i^m,
\end{equation}
where $\mathbf{x}_i$ is the node attributes of $v_i$, and $\mathbf{W}_f$, $\mathbf{W}_a$ are the learnable parameters for feature and structure generation. 
We generate the synthetic features $\mathbf{X}^g_i \in \mathbb{R}^{s \times d}$ and adjacency matrix $\mathbf{A}^g_i \in \mathbb{R}^{s \times s}$ for trigger nodes corresponding to node $v_i$. In accordance with the discrete nature of real-world graphs, we binarize $\mathbf{A}^g_i$ for the forward computation to align with the binary structure of the graph, while the continuous adjacency matrix is utilized during the gradient computation in backpropagation following \cite{Dai_2023}.

To make sure that $f_g$ can generate distribution preserving triggers, we adopt the adversarial learning strategy. Specifically, we introduce an OOD detector $f_d$ which aims to differentiate if an input  is from the original graph $\{v_i \in \mathcal{V}, Y = 1\}$ or from generated triggers $\{g_i \in \mathcal{T}_P, Y=0\}$.
Following the Generative Adversarial Network (GAN) framework in \cite{goodfellow2014generative}, 
$f_d$ refines its ability to discern between clean inputs and generated triggers by minimizing the binary classification loss.
Concurrently, the generator $f_g$ craft triggers to deceive $f_d$ by maximizing the binary classification loss.
This min-max game equips the generator with the ability to produce triggers that are indistinguishable from in-distribution data.
The min-max process is mathematically described as:
\begin{equation}
    \min_{\theta_g} \max_{\theta_d} \mathcal{L}_D = \sum_{v_i \sim \mathcal{V}_s} \log(f_d(v_i)) + \sum_{g \in \mathcal{T}_P} \log(1- f_d(g)),
\label{adver}
\end{equation}
where $\mathcal{V}_s \in \mathcal{V}$ is a selected set of representative nodes.
The reason why we select $\mathcal{V}_s$ instead of using $\mathcal{V}$ is because benchmark datasets inherently contain outlier samples at the edges of the feature distribution \cite{NEURIPS2022_acc1ec4a}. 
These outliers, when taken as inputs by the detector $f_d$, simplify the task for the trigger generator $f_g$ to deceive $f_d$ by producing triggers similar to these outliers, thereby undermining the objective of generating in-distribution triggers.
Thus, the careful choice of representative samples for training the OOD detector is critical.
To obtain $\mathcal{V}_s$, we pretrain an auto-encoder on the original graph $\mathcal{G}$ and select samples whose reconstruction losses are close to the mean loss, e.g., within one standard deviation. The details are described in Appendix \ref{representative}.
The idea behind using an autoencoder for this purpose is that it learns to identify and capture the most crucial and frequently occurring features and patterns within the data. These chosen samples can be regarded as ``typical'' based on the criteria of reconstruction loss.

\subsection{Enhancing ID Trigger Effectiveness}
Though using an OOD detector and adversarial strategy restricts $f_g$ to generating in-distribution triggers, our empirical results indicate that the attack success rate is not comparable to that achieved by GTA \cite{xi2021graph} and UGBA \cite{Dai_2023}. 
This is because the generated triggers by GTA and UGBA are outliers that deviate a lot from the original nodes, which makes it easier for the victim model to create a shortcut to associate the backdoor trigger with the target class.  
In contrast, when using ID triggers, it becomes challenging for the victim model trained on our poisoned dataset to discern a specific trigger pattern, resulting in non-activation when we attach the trigger to the target node.
To ensure the attack performance with ID triggers, we design novel modules to enhance trigger memorization by the victim model and improve attack adaptability against unseen, new targets.
Next, we give the details of each module.

\subsubsection{Enhancing Memorization of Triggers}
A key to successful graph backdoor attacks is the ability of having the victim model to correlate attached triggers with the target class. Thanks to the message-passing mechanism, trigger attributes can directly influence the attributes of the target nodes. 
Considering the diverse attributes across target nodes, we propose to encourage the generator to produce triggers that, once attached to different target nodes, can prompt the victim model to learn similar embeddings for the poisoned target nodes. Specifically, for each pair of nodes $v_i \in \mathcal{V}_P$ and $,v_j \in \mathcal{V}_P$, when attaching triggers to them, our objective is to ensure that these triggers can guide the surrogate classifier $f_s$ to learn a high cosine similarity between 
$z_s(\tilde{v}_i)$ and $z_s(\tilde{v}_j)$, where $\tilde{v}_i=a\left(\mathcal{G}_C^i, g_i\right)$ denotes $v_i$ attached with backdoor trigger $g_i$ and 
$z_s(\tilde{v}_i)$
represents the learned embedding of poisoned target node $\tilde{v}_i$ by the surrogate classifier, i.e., the last layer embedding of $f_s$ before feeding to Softmax function.
This approach ensures that the trigger attributes significantly impact the target node attributes, making them the dominant features within the embeddings of poisoned target nodes learned by the victim model. This dominance ensures that the victim model memorizes these triggers more effectively, resulting in a higher likelihood of a successful attack.

Moreover, once triggers have exerted a strong influence on the target node embeddings, enhancing the feature-level similarity between the poisoned target nodes and the nodes of the target class can further mislead the victim model into misclassifying the poisoned target nodes as belonging to the specific target class. Specifically, for a pair of nodes $v_i \in \mathcal{V}_P$ and $v_j \in \mathcal{V}_t$, where $\mathcal{V}_t \in \mathcal{V}_L$ denotes nodes from target class, our goal is to generate triggers that guide the surrogate model to learn a high cosine similarity between embeddings $z_s(\tilde{v}_i)$ and $z_s(v_j)$, while ensuring the similarity between $z_s(\tilde{v}_i)$ and $z_s(v_k)$ is lower, where $v_k \in \mathcal{V}_L\backslash\mathcal{V}_t$ belongs to non-target class. 
Combining the aforementioned two goals, we propose the following loss function for the trigger generator $f_g$:
\begin{equation}
    \begin{aligned}
        &\mathcal{L}_E=-\sum_{v_i\in \mathcal{V}_U}\sum_{v_j\in \mathcal{V}_U} S(z_s(\tilde{v}_i),z_s(\tilde{v}_j)) + \\
        &\sum_{v_i\in \mathcal{V}_U}\sum_{v_j\in \mathcal{V}_t}-\log(\frac{S(z_s(\tilde{v}_i),z_s(v_j))}{S(z_s(\tilde{v}_i),z_s(v_j))+\sum_{v_k\in {\mathcal{V}_L}\backslash\mathcal{V}_t}
      S(z_s(\tilde{v}_i),z_s(v_k))}),
    \end{aligned}
\label{enhance}
\end{equation}
where $S$ measures the cosine similarity of the embeddings.
By minimizing $\mathcal{L}_E$, the victim model trained on the poisoned dataset can better correlate the presence of triggers with the target class, ultimately leading to a higher attack success rate.

\subsubsection{Enhancing Attack Effectiveness against Unseen Targets}
\label{transferability}
To fully harness the attack budget, we propose implementing a strategy that involves assigning varying weights to accessible nodes. 
The idea is to enhance the adaptability and effectiveness of the trigger generator against new and unseen targets by prioritizing nodes that have proven to be particularly challenging to attack.
The core of our challenge lies in measuring the difficulty level of attacking each node.
To measure this, we employ the predicted probability distribution provided by the surrogate model for poisoned nodes. 
Specifically, for a given node $v_i \in \mathcal{V}_P$, $p_t^i=f_s\left(\tilde{v}_i\right)_t$ gives the probability that poisoned node $\tilde{v}_i$ is classified to the target class by surrogate model. 
A large $p_t^i$ indicates a successful attack, suggesting that the trigger generator has effectively learned to attack this target, and therefore, we assign it a smaller weight. 
Conversely, a target with a small $p_t^i$ is considered more challenging and is assigned a larger weight, directing the trigger generator to focus more on this target.
Then, we integrate this strategy into the outer loss in Eq. \eqref{eq:obj} and obtain:
\begin{equation}
    \mathcal{L}_T=\sum_{v_i \in \mathcal{V}_U}w_i \cdot l\left(f_s\left(\tilde{v}_i\right),y_t\right),
\label{secenhance}
\end{equation}
where $w_i=\exp(-p_{t}^i)$.

\subsection{Final Objective Function of DPGBA}
To ensure the effectiveness of the generated triggers, we optimize the adaptive trigger generator to successfully attack the surrogate classifier $f_s$, which is trained on the backdoored dataset. The training of the surrogate classifier is formulated as:
\begin{equation}
    \min_{\theta_s}\mathcal{L}_s(\theta_s,\theta_g)=\sum_{v_i \in \mathcal{V}_L} l\left(f_s\left(v_i\right), y_i\right)+\sum_{v_i \in \mathcal{V}_P} l\left(f_s\left(\tilde{v}_i\right), y_t\right),
\label{surro}
\end{equation}
where $\theta_s$ represents the parameters of the surrogate model $f_s$, $y_i$ is the label of labeled node $v_i\in\mathcal{V}_L$ and $y_t$ is the target class label.

Then, with $\mathcal{L}_T$ in Eq. \eqref{secenhance} aimed at misleading the surrogate model $f_s$ to predict various nodes from $\mathcal{V}$ to be $y_t$ once attached with generated triggers, 
$\mathcal{L}_D$ in Eq. \eqref{adver} constraining the in-distribution property of generated triggers, and $\mathcal{L}_E$ in Eq. \eqref{enhance} enhancing the attack performance for these in-distribution triggers, the final objective function of DPGBA is given as:
\begin{equation}
    \begin{aligned}
        \min _{\theta_g} \max_{\theta_d} \mathcal{L}   & = \mathcal{L}_T(\theta_s^{*},\theta_g)+\alpha \mathcal{L}_D(\theta_d,\theta_g) + \beta \mathcal{L}_E(\theta_s^{*},\theta_g) \\
    &s.t. \theta_s^{*} = \arg \min_{\theta_s}\mathcal{L}_s(\theta_s,\theta_g)
    \end{aligned}
\label{bilevel}
\end{equation}
where $\alpha$ and $\beta$ are scalars to control the contributions of $\mathcal{L}_D$ and $\mathcal{L}_E$, $\theta_g,\theta_s$ and $\theta_d$ represent the parameters for trigger generator $f_g$, surrogate model $f_s$ and OOD detector $f_d$, respectively.
We adopt bi-level optimization to optimize Eq. \eqref{bilevel}. Next, we give details of each optimization process.

\noindent\textbf{Lower level Optimization}
In lower-level optimization, the surrogate model $f_s$ will be trained on the backdoored dataset. We update $\theta_s$ for $N$ inner iterations with fixed $\theta_g$ to approximate $\theta_s^{*}$ as:
\begin{equation}
    \theta_s^{t+1}=\theta_s^t-\alpha_s \nabla_{\theta_s} \mathcal{L}_s(\theta_s,\theta_g),
\label{s}
\end{equation}
where $\theta_s^t$ denotes model parameters after $t$ iterations, $\alpha_s$ is the learning rate for training the surrogate model.

The OOD detector $f_d$ is optimized to enhance its capability to distinguish between clean inputs and generated triggers by maximizing $\mathcal{L}_{D}$. Similarly, we update $\theta_d$ with $K$ inner iterations with fixed $\theta_g$ to approximate $\theta_d^{*}$ as:
\begin{equation}
    \theta_d^{k+1}=\theta_d^k+\alpha_d \nabla_{\theta_d} \mathcal{L}_{D}(\theta_d,\theta_g),
\label{d}
\end{equation}
where $\theta_d^k$ denotes model parameters after $k$ iterations, $\alpha_d$ is the learning rate for training the surrogate model.

\noindent\textbf{Upper level optimization}
In the upper level optimization, the updated surrogate model parameters $\theta_s^T$ and OOD detector parameters $\theta_d^K$ are used to approximate $\theta_s^{*}$ and $\theta_d^{*}$, respectively. We then apply first-order approximation to compute gradients of $\theta_g$ by:
\begin{equation}
    \theta_g^{m+1}=\theta_g^m-\alpha_g \nabla_{\theta_g}\left(\mathcal{L}_T(\bar{\theta}_s,\theta_g)+\alpha \mathcal{L}_D(\bar{\theta}_d,\theta_g) + \beta \mathcal{L}_E(\bar{\theta}_s,\theta_g)\right),
\label{g}
\end{equation}
where $\bar{\theta}_s$ and $\bar{\theta}_d$ indicate gradient propagation stopping, $\theta_g^m$ denotes model parameters after $m$ iterations. The training algorithm of DPGBA is given in Algorithm \ref{algo}. Time complexity analysis can be found in Appendix \ref{time}.


\section{Experiments }

In this section, we will evaluate the proposed DPGBA on various datasets to answer the following research questions:
\begin{itemize}[leftmargin=*]
    \item \textbf{RQ1:} Can our framework conduct effective backdoor attacks on GNNs and simultaneously ensure in-distribution property?
    \item \textbf{RQ2:} How do the number of poisoned nodes affect the performance of backdoor attacks?
    \item \textbf{RQ3:} How do the in-distribution constraint and the enhancing attack performance module influence attack efficacy in scenarios both with and without defense mechanisms?
\end{itemize}

\subsection{Experimental settings}
\subsubsection{Datasets}
To demonstrate the effectiveness of our DPGBA, we conduct experiments on four public real-world datasets, i.e., Cora, Pubmed \cite{Sen_Namata_Bilgic_Getoor_Galligher_Eliassi-Rad_2008}, Flickr \cite{zeng2020graphsaint}, and OGB-arxiv \cite{hu2020open}, which are widely used for inductive semi-supervised node classification. Cora and Pubmed are small citation networks. Flickr is a large-scale graph that links image captions sharing the same properties. OGB-arixv is a large-scale citation network. The statistics of the datasets are summarized in Table \ref{dataset statistics}. More details of the dataset can be found in Appendix \ref{dataset}. 
\begin{table}[t]
\centering
\caption{Dataset Statistics}
\vskip -1em
\begin{tabular}{lllll}
\hline
\label{dataset statistics}
\textbf{Datasets} & \textbf{\#Nodes} & \textbf{\#Edges} & \textbf{\#Features} & \textbf{\#Classes} \\ \hline
Cora       & 2,708   & 5,429    & 1,443    & 7 \\
Pubmed     & 19,717  & 44,338   & 500      & 3 \\
Flickr     & 89,250  & 899,756  & 500      & 7 \\
OGB-arxiv  & 169,343 & 1,166,243 & 128     & 40 \\ \hline
\end{tabular}
\end{table}

\subsubsection{Compared Methods}
We compare DPGBA with representative and state-of-the-art graph backdoor attack methods, including UGBA \cite{Dai_2023}, GTA \cite{xi2021graph}, SBA-Samp \cite{zhang2021backdoor} and its variant SBA-Gen.
More details of these compared methods can be found in Appendix \ref{compared}. For a fair comparison, hyperparameters of all the attack methods are tuned based on the performance of the validation set.

\begin{table*}[htbp]
\centering
\caption{Backdoor attack results (ASR (\%) | Clean Accuracy (\%)). Only clean accuracy is reported for clean graphs. The best results are marked with boldface.}
\vskip -1em
\begin{tabular}{cccccccc}
\hline
\textbf{Datasets} & \textbf{Defense} & \textbf{Clean Graph} & \textbf{SBA-Samp} & \textbf{SBA-Gen} & \textbf{GTA} & \textbf{UGBA} & \textbf{DPGBA} \\ \hline
\multirow{2}*{Cora} & None       &  83.9     & 35.2 | 83.5 & 45.3 | 83.2 & 91.6 | \textbf{83.6} & 95.1 | 83.5 & \textbf{96.7} | \textbf{83.6} \\ 
     & OD         &  83.8     & 34.7 | 83.0 & 44.1 | \textbf{83.6} & 0.00 | 83.4 & 0.00 | \textbf{83.6} & \textbf{93.9} | 83.5 \\ \hline
\multirow{2}*{Pubmed} & None     &  85.1     & 33.8 | 84.7 & 34.4 | 84.6 & 88.1 | 84.9 & 92.5 | \textbf{85.2} & \textbf{92.6} | 85.1 \\ 
       & OD       &  85.0     & 33.3 | 84.9 & 33.5 | 84.6 & 0.00 | 84.7 & 0.00 | 85.0 & \textbf{91.8} | \textbf{85.1} \\ \hline
\multirow{2}*{Flickr} & None     &  46.0     & 0.00 | \textbf{46.2} & 0.00 | 45.8 & 88.6 | 45.0 & 94.9 | 45.4 & \textbf{96.4} | 45.9 \\ 
       & OD       &  46.2     & 0.00 | \textbf{45.9} & 0.00 | 45.6 & 0.00 | 45.1 & 0.00 | 45.4 & \textbf{94.8} | 45.8 \\ \hline
\multirow{2}*{OGB-arxiv} & None  &  65.9     & 36.0 | 65.8 & 43.4 | \textbf{65.9} & 92.5 | 65.8 & \textbf{98.2} | 65.3 & 95.1 | 65.6 \\ 
          & OD    &  65.8     &  35.0 | \textbf{65.6} & 42.3 | 65.5 & 0.00 | 64.9 & 0.00 | 64.5 & \textbf{92.4} | 65.4 \\ \hline

\end{tabular}
\label{maintable}
\end{table*}

\subsubsection{Evaluation Protocol} 
\label{protocol}
Following the evaluation protocol in UGBA \cite{Dai_2023}, we conduct experiments on the inductive node classification task.
In this setup, attackers do not have access to test node during trigger generator training.
We randomly exclude $20\%$ of nodes from the original dataset, denoted as $\mathcal{V}_T$, using half as targets for assessing attack effectiveness and the other half as clean test nodes for evaluating the accuracy of models under attack on normal samples.
The training graph $\mathcal{G}$ consists of the remaining $80\%$ of nodes, with the labeled node set and validation set each containing $10\%$ of nodes. We measure backdoor attack performance using the average success rate (ASR) on target nodes and clean accuracy on clean test nodes. A two-layer GCN acts as the surrogate model for all attack strategies. To evaluate the transferability of backdoor attacks, we target GNNs with different architectures—GCN, GraphSage, and GAT. We conduct experiments on each GNN architecture five times and report the average ASR and clean accuracy from the total of 15 experiments.  
The attack budget $\Delta_P$ on size of poisoned nodes $\mathcal{V}_P$ is set as 10, 40, 160, and 565 for Cora, Pubmed, Flickr, and OGB-arxiv, respectively.
The number of nodes in the trigger size is limited to 3 for all experiments.
Our DPGBA deploys a 2-layer GCN as the surrogate model. A 2-layer MLP is used as the trigger generator.
More details of the hyperparameter setting can be found in Appendix \ref{impledetail}.

For the defense strategy \textbf{OD}, in line with the in-distribution constraint outlined in Sec. \ref{idcons}, we use DOMINANT \cite{ding2019deep} as $f_o$ and train it on the poisoned graph $\mathcal{G}_B$ with triggers attached to nodes in $\mathcal{V}_P$. The threshold $\tau$, as specified in Eq. \eqref{thres}, is set such that data points with a reconstruction loss greater than $\tau$ comprise $3\%$ of the dataset. The remaining $97\%$ of the data points have a reconstruction loss at or below $\tau$. 
Before training the surrogate model $f_s$ on the poisoned graph $\mathcal{G}_B$, we prune those nodes with reconstruction loss above $\tau$. Once $f_s$ is trained, $f_a$ and $\tau$ are fixed for the testing phase to perform inference on test nodes in $\mathcal{V}_T$ and the associated generated triggers. Nodes with a reconstruction loss above $\tau$ are pruned. 



\subsection{Backdoor Attack Performance}
To answer \textbf{RQ1}, we evaluate DPGBA against baseline methods on four real-world graphs, considering scenarios with and without the OD defense strategy as outlined in Sec. \ref{protocol}. We report the average results in backdooring three different GNN architectures in Tab. \ref{maintable}.
Detailed results for each architecture are provided in Tab. \ref{GCN}~--~\ref{GAT}
in Appendix \ref{exontrans}. From the table, we make the following observations:
\begin{itemize}[leftmargin=*]
    \item When no backdoor defense strategy is applied, DPGBA shows a comparable or slightly better ASR than leading baselines such as GTA and UGBA, while SBA-Samp and its variant, SBA-Gen, consistently achieve lower ASRs. This indicates the effectiveness of our modules in enhancing the influence of triggers on the target nodes.
    Regarding clean accuracy, our framework consistently demonstrates comparable results with all the baselines.
    \item When applying a simple outlier detection defense, triggers generated by GTA and UGBA are removed, but DPGBA still achieves over $90\%$ ASR. This demonstrates that our DPGBA effectively generates imperceptible ID triggers that can successfully bypass commonly used outlier detection methods in real applications.
    \item Though we employ GCN as the surrogate model during training, the generated triggers consistently achieve high ASR across three different GNN architectures, as shown in Tab. \ref{GCN}~--~\ref{GAT}. This indicates the transferability of the trigger generator within our framework.
\end{itemize}

\begin{figure}[t]
    \centering
    \includegraphics[width=.42\textwidth]{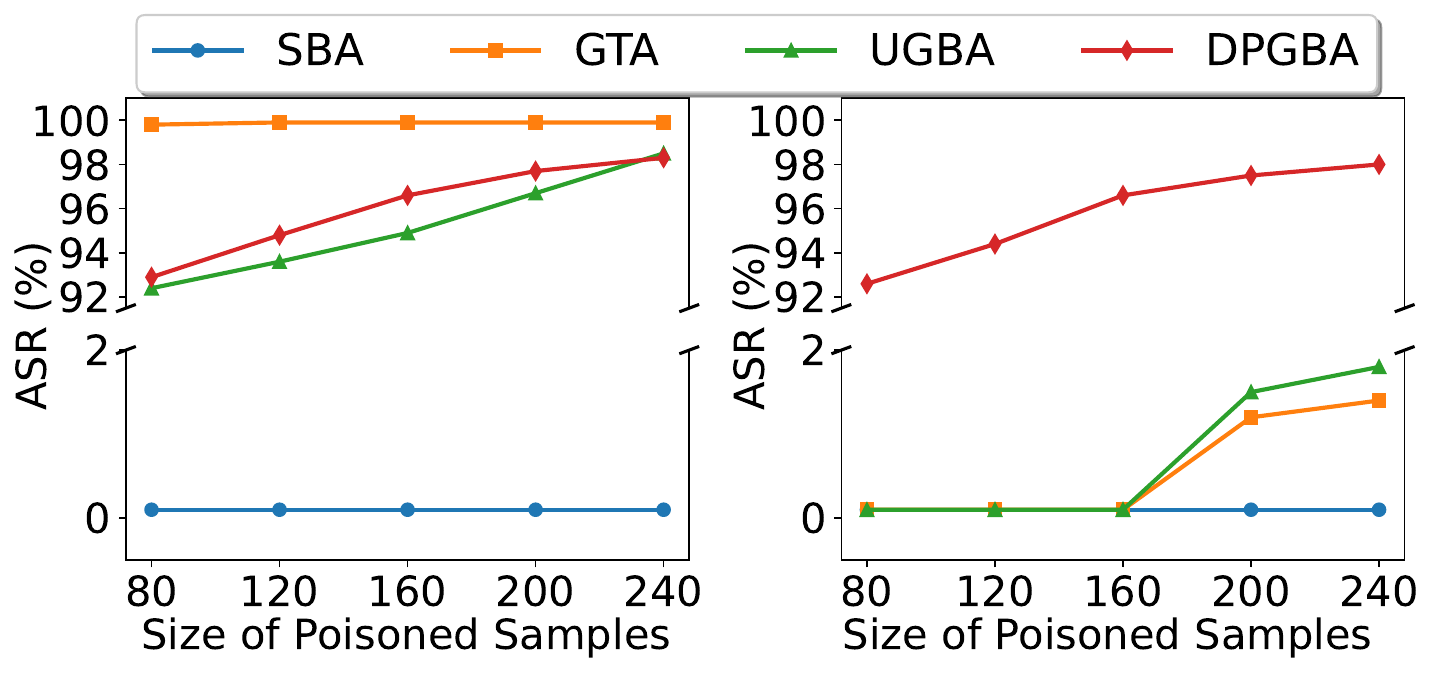}
    \begin{minipage}{0.25\textwidth}
        \centering
        \textbf{(a) No defense}
        \vspace{1mm}
    \end{minipage}%
    \begin{minipage}{0.25\textwidth}
        \centering
        \textbf{(b) OD}
        \vspace{1mm}
    \end{minipage}
    \vskip -1em
    \caption{Impacts of sizes of poisoned nodes on Flicker.
    }
    \label{SPS}
\end{figure}

\subsection{Impact of the Size of Poisoned Nodes}
To answer \textbf{RQ2}, 
we conduct experiemnts to explore the attack performance of DPGBA given different budgests in the size of poisoned nodes. Specifically, we vary the size of poisoned samples as $\{40,80,120,160,200\}$. The other settings are the same as the evaluation protocol in Sec. \ref{protocol}.
Hyperparameters are selected with the same process as described in Appendix \ref{impledetail}.
Fig. \ref{SPS} shows the results on Flicker dataset.
We have similar observations on other datasets. 
We only report the attack success rate as we did not observe any significant change in clean accuracy for all the baselines and our DPGBA.
From Fig. \ref{SPS}, we can observe that:
\begin{itemize}[leftmargin=*]
    \item The attack success rate of UGBA \cite{Dai_2023} and DPGBA consistently rises as the number of poisoned samples increases in (a), which aligns with our expectation. Our method maintains a comparable ASR when no defense is applied, highlighting the effectiveness of our attack performance enhancement module.
    \item When OD defense is applied on the backdoor attacks in (b), our DPGBA still achieve promising performances. In contrast, all the baseline methods achieve an almost $0\%$ ASR, as anticipated. That is because our method can generate trigger nodes with in-distribution property.
    
\end{itemize}

\subsection{In-distribution Property Analysis}
\label{indisana}
\begin{figure}[t]
    \centering
    \includegraphics[width=.46\textwidth]{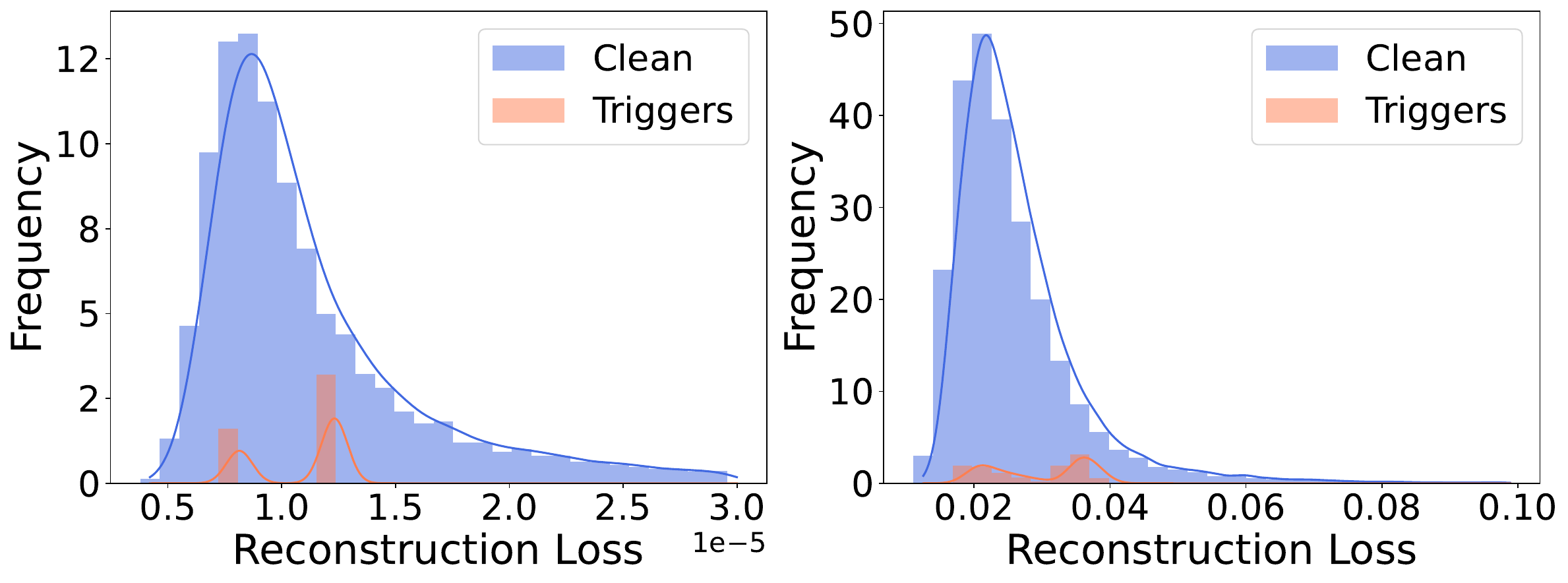}
    \begin{minipage}[t][-0.8cm][c]{.5\textwidth}
        \centering
        \caption*{(a) Flicker\ \ \ \ \ \ \ \ \ \ \ \ \ \ \ \ \ \ \ \ \ \ \ \ \ \ \ \ \ \ \ (b) OGB-arxiv}
    \end{minipage}
    \vspace{-8mm}
    \caption{Reconstruction loss distributions on Flicker and OGB-arxiv} 
    \label{ID}
\end{figure}

In this subsection, to further demonstrate the in-distribution property of triggers generated by our framework, we first conduct backdoor attacks on Flicker and OGB-arxiv datasets, then apply the outlier detection method on the poisoned graph, and finally show the reconstruction loss for both clean data and generated triggers. 
The histograms of the reconstruction loss are plotted in Fig. \ref{ID}. 
From the figure, we observe that the reconstruction loss of the generated triggers closely aligns with the mean of the distribution of reconstruction losses for clean inputs. This alignment can be attributed to the selection of representative samples $\mathcal{V}_S$ for the OOD detector $f_d$ and adversarial learning to make the trigger in-distribution. Additional experiments in Appendix \ref{againstvarious} demonstrate the efficacy of our generated triggers in bypassing various advanced graph outlier detection methods.



\subsection{Ablation Studies}
\begin{figure}[t]
    \centering
    \includegraphics[width=.46\textwidth]{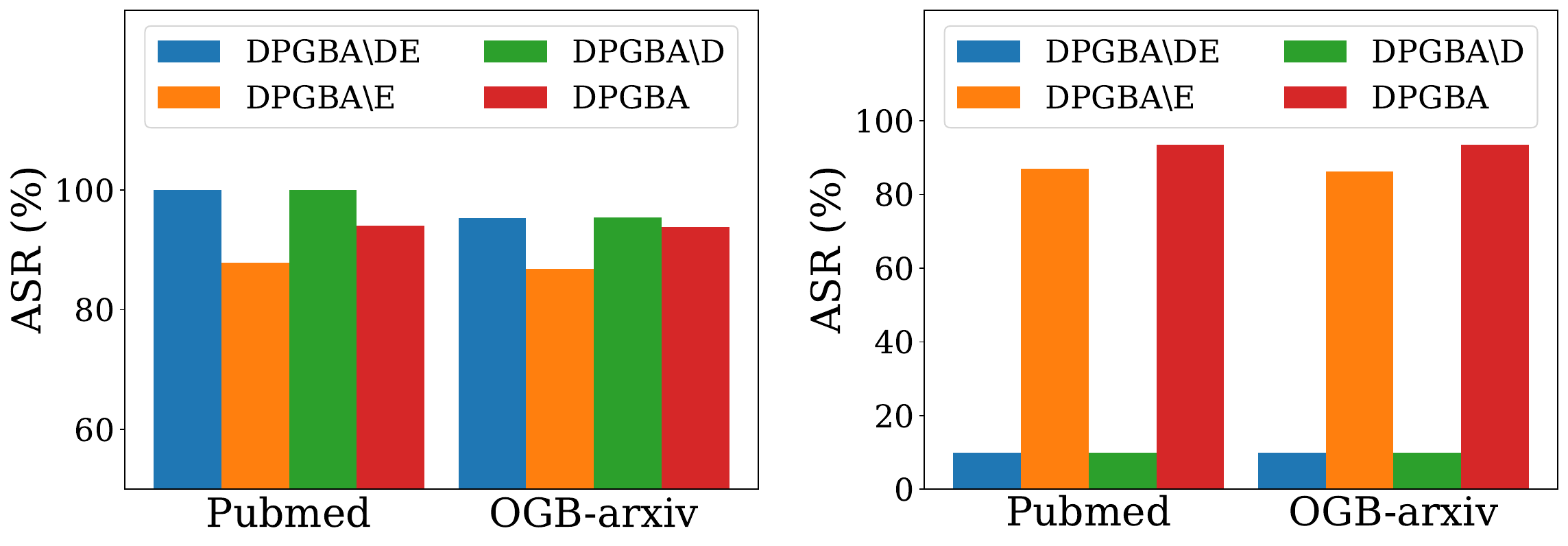}
    \begin{minipage}{0.25\textwidth}
        \centering
        \textbf{(a) No defense}
        \vspace{1mm}
    \end{minipage}%
    \begin{minipage}{0.25\textwidth}
        \centering
        \textbf{(b) OD}
        \vspace{1mm}
    \end{minipage}
    
    \vspace{-4mm}
    \caption{Ablation studies on Pubmed and OGB-arxiv} 
    \label{ablation}
\end{figure}

To answer \textbf{RQ3},
we conduct ablation studies to explore the effects of the ID constraint and the enhancing triggers attack performance module.
To demonstrate the effectiveness of the ID constraint module, we set $\alpha=0$ and obtain a variant named as DPGBA\textbackslash D. To show the benefits brought by our enhancing attack performance module, we train a variant DPGBA\textbackslash E which set the $\beta$ as 0.  We also implement a variant of our model by removing both ID constraint and enhancing attack performance module, which is named as DPGBA\textbackslash DE.
The average results and standard deviations on Pubmed and OGB-arxiv are shown in Fig. \ref{ablation}.
All the settings of evaluation follow the description in Sec. \ref{protocol}.
And the hyperparameters of the variants are also tuned based on the validation set for fair comparison. From Fig. \ref{ablation}, we observe that: (\textbf{i}) When no defense method is applied, DPGBA demonstrates a comparable attack performance, despite DPGBA\textbackslash DE and DPGBA\textbackslash D taking shortcuts to generate outlier triggers. However, when the OD defense method is employed, DPGBA still exhibits a high ASR, while triggers generated by DPGBA\textbackslash DE and DPGBA\textbackslash D are almost all eliminated. This observation indicates the effectiveness of the proposed ID constraint module in generating ID triggers; and (\textbf{ii}) Compared to DPGBA\textbackslash E, DPGBA achieves superior attack performance under both defense settings, which shows the effectiveness of our enhancing attack performance module.

\subsection{Hyper-parameter Sensitivity Analysis}
\begin{figure}[htbp]
    \centering
    \includegraphics[width=.46\textwidth]{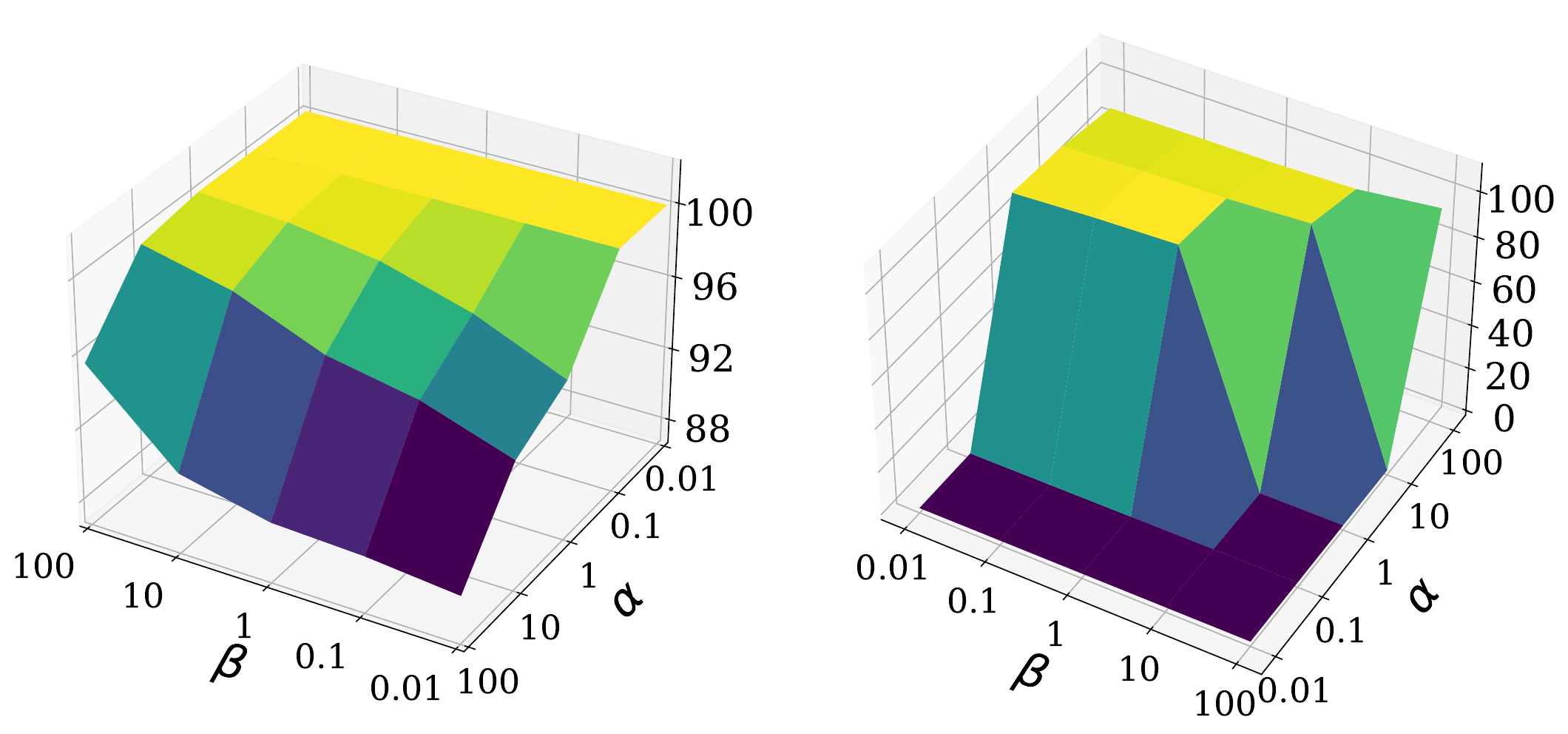}
    \begin{minipage}{0.25\textwidth}
        \centering
        \textbf{(a) No defense}
        \vspace{1mm}
    \end{minipage}%
    \begin{minipage}{0.25\textwidth}
        \centering
        \textbf{(b) OD}
        \vspace{1mm}
    \end{minipage}
    \vskip -1em
    \caption{Hyperparameter Sensitivity Analysis} 
    \label{hyper}
\end{figure}

In this subsection, we further investigate how the hyperparameter $\alpha$ and $\beta$ affect the performance of DPGBA, where $\alpha$ and $\beta$ control the weight of ID constraint and enhancing attack performance module, respectively.
To explore the effects of $\alpha$ and $\beta$, we vary the values of $\alpha$ and $\beta$ as $\left \{ 0.01,0.1,1,10,100 \right \} $
for Flicker dataset.
We report the attack success rate (ASR) of attacking in both no defense and OD defense settings in Fig. \ref{hyper}. The test model is fixed as GCN. 
We observe that 
(\textbf{i}) In the absence of defense strategies, increasing $\beta$ improves attack effectiveness, while higher $\alpha$ values lead to reduced attack performance.
(\textbf{ii}) With outlier detection method deployed, to preserve the in-distribution characteristic of generated triggers and ensure a high attack success rate, it is recommended to set $\alpha \ge 1$ and adjust $\beta$ accordingly as $\alpha$ increases, ensuring $\beta$ remains closely aligned with $\alpha$. This observation eases hyperparameter tuning.

\section{Conclusion}
In this paper, we empirically verify that existing backdoor attack methods on graph suffer from either a low attack success rate or outlier issues, which can be leveraged by outlier detection methods to identify and remove those triggers, thus significantly degrading their attack performance. To address these problems, we study a novel problem of conducting effective distribution-preserving graph backdoor attacks. Specifically, an Out-Of-Distribution (OOD) detector, in conjunction with an adversarial learning strategy, is implemented to constrain the in-distribution property of generated triggers. Additionally, a novel module is proposed to guide the victim model trained on the poisoned dataset to better correlate the presence of triggers with the target class.
Extensive experiments on large-scale datasets demonstrate that our proposed method can effectively bypass commonly used outlier detection methods in real-world applications while achieving a high attack success rate in backdooring various target GNN models.

\section*{ACKNOWLEDGMENTS}
This material is based upon work supported by, or in part by, Army Research Office (ARO) under grant number W911NF-21-1- 0198, Department of Homeland Security (DHS) CINA under grant number E205949D, and Cisco Faculty Research Award.


\bibliographystyle{ACM-Reference-Format}
\bibliography{ref}

\appendix

\begin{algorithm}
\caption{Algorithm of DPGBA}
\begin{algorithmic}[1]
\REQUIRE Graph $\mathcal{G} = (\mathcal{V}, \mathcal{E}, \mathbf{X}), \mathcal{Y}_L$, $\beta$, $T$
\ENSURE Backdoored dataset $\mathcal{G}_B$, trigger generator $f_g$
\STATE Initialize $\mathcal{G}_B = \mathcal{G}$;
\STATE Randomly initialize $\theta_s$, $\theta_d$ and $\theta_g$ for $f_s$, $f_d$ and $f_g$;
\STATE Randomly select poisoned nodes $\mathcal{V}_P$;
\STATE Assign class $t$ as labels of $\mathcal{V}_P$;
\WHILE{not converged yet}
    \FOR{$t = 1, 2, \ldots, N$}
        \STATE Update $\theta_s$ by descent on $\nabla_{\theta_s} \mathcal{L}_s$ based on Eq. \eqref{s};
    \ENDFOR
    \FOR{$t = 1, 2, \ldots, K$}
        \STATE Update $\theta_d$ by descent on $\nabla_{\theta_d} \mathcal{L}_{D}$ based on Eq. \eqref{d};
    \ENDFOR
    \STATE Update $\theta_g$ by descent on $\nabla_{\theta_g} \mathcal{L}_T+\alpha \mathcal{L}_D + \beta \mathcal{L}_E$ based on Eq. \eqref{g};
\ENDWHILE
\FOR{each $v_i \in \mathcal{V}_P$}
    \STATE Generate the trigger $g_i$ for $v_i$ by using $f_g$;
    \STATE Update $\mathcal{G}_B$ based on $a(\mathcal{G}_{B}^i, g_i)$;
\ENDFOR
\RETURN $\mathcal{G}_B$ and $f_g$;
\end{algorithmic}
\label{algo}
\end{algorithm}
\section{TRAINING ALGORITHM}
The DPGBA algorithm is detailed in Algorithm \ref{algo}. Initially, we identify the poisoned nodes $\mathcal{V}_P$ and label them with the target class $y_t$ (lines 3-4). From lines 5-13, the trigger generator $f_g$ is trained to both attack the surrogate model $f_s$ and deceive the OOD detector $f_d$, utilizing a bi-level optimization approach. Specifically, in the lower level, we update the surrogate model (lines 6-8) and the OOD detector (lines 9-11) through gradient descent on $\theta_s$ and $\theta_d$, respectively, guided by Eq. \eqref{s} for $f_s$ and Eq. \eqref{d} for $f_d$. In the upper level, the generator $f_g$ is updated (line 12) by applying gradient descent on $\theta_g$, as outlined in Eq. \eqref{g}.
After that, from line 14 to 17, we use the well-trained $f_g$ to generate a trigger $g_i$ for each poisoned node $v_i \in \mathcal{V}_P$ and attach $g_i$ with $v_i$ to obtain the poisoned graph $\mathcal{G}_B$.
\section{REPRESENTATIVE NODES SELECTION}
\label{representative}
For selecting representative nodes $\mathcal{V}_s$ from a clean graph $\mathcal{G}$, we use the outlier detection method outlined in DOMINANT \cite{ding2019deep}. This approach is first applied to $\mathcal{G}$ to determine the mean $\mu$ and standard deviation $\delta$ of the reconstruction losses. Representative nodes are then selected based on their reconstruction loss $d$, ensuring that $d < \mu + \gamma\delta$. The parameter $\gamma$ is set to 1 for the Cora dataset and adjusted to 0.01 for the Pubmed, Flicker, and OGB-arxiv datasets.

\begin{table*}[htbp]
\centering
\caption{Results of backdooring GCN (ASR (\%) | Clean Accuracy (\%)). Only clean accuracy is reported for clean graph.}
\label{GCN}
\vskip -1em
\small
\begin{tabular}{llllllll}
\hline
\textbf{Datasets} & \textbf{Defense} & \textbf{Clean Graph} & \textbf{SBA-Samp} & \textbf{SBA-Gen} & \textbf{GTA} & \textbf{UGBA} & \textbf{DPGBA}  \\ \hline
\multirow{2}*{Cora}              & None             &      83.6            & 33.8±5.2 | 83.4±0.8 & 48.3±8.2 | 83.3±0.5 & 94.6±1.6 | 83.6±1.0 & 98.3±0.1 | 83.5±0.8 & 97.7±1.4 | 83.3±0.8 \\
                  & OD               &      83.4           & 33.4±4.9 | 82.9±0.6& 47.8±9.3 | 83.7±0.6  & 0±0.0 | 83.4±0.7 & 0±0.0 | 83.7±0.5 &94.4±1.1 | 83.5±1.0 \\ \hline
\multirow{2}*{Pubmed}            & None             &      85.1            & 36.5±11.4 | 85.1±0.2 & 36.1±3.7 | 85.0±0.1  & 88.8±1.7 | 85.1±0.2 & 93.1±1.3 | 85.1±0.2 & 92.3±1.8 | 85.0±0.2\\
                  & OD               &      85.1            & 35.8±12.1 | 85.2±0.1  & 35.5±3.2 | 85.2±0.1 & 0±0.0 | 85.3±0.3& 0±0.0 | 85.2±0.1 & 91.2±1.2 | 85.1±0.2\\ \hline
\multirow{2}*{Flickr}            & None             &      46.2           & 0±0.0 | 45.5±0.2    & 0±0.0 | 45.5±0.1    & 99.9±0.1 | 45.0±0.3 & 96.9±2.3 | 44.8±0.4 & 98.8±1.6 | 46.4±0.4\\
                  & OD               &      46.0           & 0±0.0 | 45.8±0.4 & 0±0.0 | 45.3±0.2  &0±0.0 | 45.3±0.4 & 0±0.0 | 44.4±0.3&96.0±1.7 | 45.9±0.2 \\ \hline
\multirow{2}*{OGB-arxiv}         & None             &   66.2   & 35.2±5.7 | 65.9±0.1   & 48.5±4.2 | 65.9±0.1     & 83.6±2.8 | 65.3±0.3   & 99.4±0.1 | 65.3±0.5 & 95.6±0.8 | 65.8±0.4 \\
                  & OD               &   65.9  & 34.6±6.6 | 65.6±0.2   & 47.2±4.5 | 65.4±0.3    & 0±0.0 | 64.7±0.3 & 0±0.0 | 65.4±0.5&93.2±1.0 | 65.5±0.3\\ \hline
\end{tabular}
\end{table*}

\begin{table*}[htbp]
\centering
\caption{Results of backdooring GraphSage (ASR (\%) | Clean Accuracy (\%)). Only clean accuracy is reported for clean graph.}
\label{SAGE}
\small
\vskip -1em
\begin{tabular}{llllllll}
\hline
\textbf{Datasets} & \textbf{Defense} & \textbf{Clean Graph} & \textbf{SBA-Samp} & \textbf{SBA-Gen} & \textbf{GTA} & \textbf{UGBA} & \textbf{DPGBA}  \\ \hline
\multirow{2}*{Cora}              & None       &   83.8     &  34.2±4.0 | 83.0±1.5 & 40.4±5.6 | 82.7±1.2  & 96.0±3.3 | 83.3±1.2  & 92.7±2.1 | 83.6±1.4 & 95.3±1.5 | 83.7±1.2 \\
                  & OD         &   83.6    & 33.9±3.3 | 82.5±1.1 & 38.3±5.3 | 83.1±1.3  & 0±0.0 | 82.8±1.1 & 0±0.0 | 83.3±1.7&91.2±1.1 | 83.5±0.8 \\ \hline
\multirow{2}*{Pubmed}            & None       &   84.9    &   38.0±3.8 | 84.8±0.3 & 40.0±4.2 | 84.9±0.2  & 89.0±6.4 | 84.9±0.2  & 90.2±1.0 | 85.1±0.1 &91.8±1.3 | 85.1±0.4 \\
                  & OD         &   85.0    & 37.8±3.3 | 85.4±0.2 &39.2±5.3 | 84.8±0.2  &0±0.0 | 85.1±0.5  &0±0.0 | 85.0±0.1  &91.0±1.2 | 85.2±0.3 \\ \hline
\multirow{2}*{Flickr}            & None       &   46.0    &   0±0.0 | 45.5±0.1 & 0±0.0 | 45.4±0.1  & 99.7±0.2 | 46.0±0.3  & 91.5±2.1 | 45.7±0.3 & 94.8±1.8 | 45.6±0.2 \\
                  & OD         &   46.3    & 0±0.0 | 45.0±0.3 & 0±0.0 | 45.1±0.1 & 0±0.0 | 46.3±0.1 &0±0.0 | 46.0±0.2 &93.3±2.4 | 45.7±0.3 \\ \hline
\multirow{2}*{OGB-arxiv}         & None       &   65.8    & 33.0±5.6 | 66.1±0.4  & 38.7±1.9 | 66.1±0.3  & 99.6±0.3 | 64.4±0.4    & 97.7±0.1 | 65.5±0.1 & 94.2±0.8 | 65.8±0.6 \\
                  & OD         &   66.1    & 32.2±5.9 | 65.6±0.6 & 37.6±2.3 | 65.8±0.2 & 0±0.0 | 64.8±0.3 &0±0.0 | 65.6±0.2 &91.8±0.6 | 65.4±0.4\\ \hline
\end{tabular}
\end{table*}

\begin{table*}[htbp]
\centering
\caption{Results of backdooring GAT (ASR (\%) | Clean Accuracy (\%)). Only clean accuracy is reported for clean graph.}
\label{GAT}
\vskip -1em
\small
\begin{tabular}{llllllll}
\hline
\textbf{Datasets} & \textbf{Defense} & \textbf{Clean Graph} & \textbf{SBA-Samp} & \textbf{SBA-Gen} & \textbf{GTA} & \textbf{UGBA} & \textbf{DPGBA}  \\ \hline
\multirow{2}*{Cora}              & None       &  84.3   &   37.5±8.7 | 84.0±1.3 & 47.1±18.0 | 83.7±1.1  & 84.1±3.8 | 83.9±0.9 & 94.3±1.4 | 83.3±0.7 & 97.1±1.7 | 83.7±1.0 \\
                  & OD         &  84.4   &    36.8±7.7 | 83.6±2.2 & 46.3±17.6 | 83.9±1.3 & 0±0.0 | 84.1±0.6 &0±0.0 | 83.7±0.9 &96.0±2.0 | 83.6±0.8 \\ \hline
\multirow{2}*{Pubmed}            & None       &  85.2   &   26.9±4.5 | 84.1±0.3 & 27.1±3.8 | 83.9±0.2  & 86.4±2.6 | 83.8±0.2 & 94.2±1.5 | 85.4±0.1 & 93.8±2.6 | 85.1±0.1\\
                  & OD         &  84.9   &   26.3±3.7 | 84.2±0.4 & 25.8±4.4 | 83.6±0.2  &0±0.0 | 83.6±0.5 &0±0.0 | 84.8±0.3 &93.3±2.1 | 85.0±0.2  \\ \hline
\multirow{2}*{Flickr}            & None       &  46.7   &    0±0.0 | 46.5±0.2 & 0±0.0 | 46.6±0.4   & 66.2±34.9 | 44.0±0.6  & 96.2±4.2 | 45.6±0.3 & 95.7±4.4 | 45.6±0.2\\
                  & OD         &  46.4   & 0±0.0 | 46.9±0.4& 0±0.0 | 46.4±0.7 &0±0.0 | 43.7±0.4 & 0±0.0 | 45.8±0.4& 95.1±3.6 | 45.8±0.3\\ \hline
\multirow{2}*{OGB-arxiv}         & None       &  65.6   & 39.7±7.2 | 65.3±0.3    & 43.0±10.4 | 65.6±0.4 &     94.3±2.5 | 64.8±0.1  & 97.6±0.1 | 65.1±0.2 & 95.4±1.3 | 65.2±0.2\\
                  & OD         &  65.3   & 38.3±6.1 | 65.5±0.5  &  42.1±11.3 | 65.3±0.3  & 0±0.0 | 65.1±0.5 & 0±0.0 | 65.5±0.4 & 92.1±0.9 | 65.4±0.3\\ \hline
\end{tabular}
\end{table*}

\section{DATASETS DETAILS}
\label{dataset}
\begin{itemize}[leftmargin=*]
    \item \textbf{Cora} and \textbf{PubMed} \cite{yang2016revisiting}: They are citation networks where nodes denote papers, and edges depict citation relationships. In Cora and CiteSeer, each node is described using a binary word vector, indicating the presence or absence of a corresponding word from a predefined dictionary. In contrast, PubMed employs a TF/IDF weighted word vector for each node. For all three datasets, nodes are categorized based on their respective research areas.
    \item \textbf{Flicker} \cite{zeng2020graphsaint}: In this graph, each node symbolizes an individual image uploaded to Flickr. An edge is established between the nodes of two images if they share certain attributes, such as geographic location, gallery, or user comments. The node features are represented by a 500-dimensional bag-of-word model provided by NUS-wide. Regarding labels, we examined the 81 tags assigned to each image and manually consolidated them into 7 distinct classes, with each image falling into one of these categories.
    \item \textbf{OGB-arxiv} \cite{hu2020open}: It is a citation network encompassing all Computer Science arXiv papers cataloged in the Microsoft Academic Graph. Each node is characterized by a 128-dimensional feature vector, which is derived by averaging the skipgram word embeddings present in its title and abstract. Additionally, the nodes are categorized based on their respective research areas.
\end{itemize}

\section{ADDITIONAL RELATED WORKS}
\label{odrelated}
Graph outlier detection is a critical task in machine learning, involving the identification of anomalous nodes within a graph. Unlike traditional outlier detection on tabular or time-series data, graph outlier detection presents unique challenges due to the rich information inherent in graph structures and the computational complexity of training with complex machine learning models. The emergence of deep learning techniques has revolutionized outlier detection, shifting from traditional methods to neural network approaches \cite{ma2021comprehensive}. One popular neural network architecture for this task is the autoencoder (AE) \cite{kingma2013auto}, which learns to reconstruct the original data and identifies outliers based on reconstruction errors. This unsupervised learning approach makes AEs effective for detecting outliers without the need for labeled data. Furthermore, graph neural networks (GNNs) have demonstrated superior performance in capturing complex patterns within graph data, considering both node attributes and graph structure. GNNs encode representations for each node, enabling effective outlier detection. Notably, GNNs can be combined with AEs \cite{ding2019deep, fan2020anomalydae, kipf2016variational,  dai2022graphaugmented, bandyopadhyay2019outlier, xu2022contrastive}, leveraging the strengths of both approaches for more robust outlier detection in graph data. 

\section{EXPERIMENTS ON ATTACK TRANSFERABILITY}
\label{exontrans}

To demonstrate the transferability of our trigger generator in attacking various GNN architectures, we employ GCN as the surrogate model and evaluate the ASR and clean accuracy when attacking GCN, GraphSage \cite{hamilton2017inductive} and GAT \cite{veličković2018graph}, respectively. The results are presented in Tab. \ref{GCN}~--~\ref{GAT}. From the tables, we observe that our DPGBA consistently achieves a high attack success rate while maintaining the clean accuracy across different target models and various defense settings. 
This indicates the adaptability and transferability of our framework, enhancing its practical value in real-world applications.

\section{AGAINST VARIOUS OUTLIER DETECTION METHODS}
\label{againstvarious}
To further demonstrate the in-distribution property of the triggers generated by our DPGBA, we adopt various state-of-the-art graph outlier detection methods, including DOMINANT \cite{ding2019deep}, DONE \cite{bandyopadhyay2019outlier} and its variant AdONE, AnomalyDAE \cite{fan2020anomalydae}, GAAN \cite{chen2020generative} and CONAD \cite{xu2022contrastive}, as defense mechanisms and conduct backdoor attacks on four datasets. The other settings are the same as the evaluation protocol in Sec. \ref{protocol}. The results of ASR are reported in Tab. \ref{againstvarioustable}. From the table, we observe that DPGBA consistently exhibits its capability to evade various graph outlier detection methods and maintain a high attack success rate. This consistency underscores the practical application value of DPGBA in real-world scenarios.

\begin{table}[htbp]
\centering
\caption{Backdoor attack results against various graph outlier detection methods}
\vskip -1em
\begin{tabular}{ccccc}
\hline
\textbf{Defense} & \textbf{Cora} & \textbf{Pubmed} & \textbf{Flicker} & \textbf{OGB-arxiv} \\ \hline

None &  97.7    & 92.3  & 98.8  & 95.6   \\
DOMINANT &  94.4   & 91.2  & 96.0  & 93.2   \\
DONE   & 94.3  &  90.9  & 97.3  & 94.4   \\
AdONE  &  95.7    & 92.0    & 98.0    & 93.4      \\
AnomalyDAE &  96.4    & 91.7  &    97.4  & 95.1  \\
GAAN  &    96.8   & 91.8  & 98.6  & 94.9     \\
CONAD  &  96.6  & 91.3  &  98.5  &  94.7  \\ \hline
\label{againstvarioustable}
\end{tabular}
\end{table}





\section{ADDITIONAL EXPERIMENTS}

To demonstrate the robust adaptability of the trigger generator within DPGBA, we compare DPGBA with UGBA \cite{Dai_2023} using the defense strategy \textbf{Prune} proposed in \cite{Dai_2023}, which involves removing edges connecting nodes with low cosine similarity. All experimental configurations adhere to the evaluation protocol outlined in Section \ref{protocol}. \
Following \cite{Dai_2023}, we incorporate the unnoticeable loss proposed in \cite{Dai_2023} to ensure that generated triggers exhibit high cosine similarity to target nodes.
We set the pruning threshold to exclude approximately $10\%$ of dissimilar edges. Table \ref{additional} presents the results of ASR and clean accuracy. 
From the table, we observe that DPGBA consistently exhibits comparable ASR and slightly higher clean accuracy compared to UGBA across four datasets. Notably, generated triggers in DPGBA maintain in-distribution property, whereas UGBA fails to evade detection by outlier detection methods.
These findings indicate the superior performance and robustness of DPGBA in diverse settings.

\begin{table}[htbp]
\centering
\caption{Backdoor attack results (ASR (\%) | Clean Accuracy (\%)). Only clean accuracy is reported for clean graphs.}
\vskip -1em
\begin{tabular}{ccccc}
\hline
\textbf{Datasets} & \textbf{Defense} & \textbf{Clean Graph} & \textbf{UGBA} & \textbf{DPGBA} \\ \hline
\multirow{2}*{Cora} 
& OD &  83.4   & \ \ 0.0 | 83.7 &94.4 | 83.5   \\ 
     & Prune  &  83.6   &  95.9 | 82.5 &  91.8 | 85.2  \\ \hline
\multirow{2}*{Pubmed} 
& OD & 85.1  & \ \ 0.0 | 85.2 & 91.2 | 85.1\\
       & Prune &   85.1    &  89.1 | 85.4   &   88.6 | 85.1    \\ \hline
\multirow{2}*{Flickr} 
& OD & 46.2  &  \ \ 0.0 | 44.4 & 96.0 | 45.9 \\
       & Prune & 45.3    &   99.7 | 41.7 &94.7 | 45.9  \\ \hline

\multirow{2}*{OGB-arxiv} 
& OD & 65.8 & \ \ 0.0 | 65.4 &93.2 | 65.5\\ 
       & Prune  &  66.3    &   93.4 | 63.0 & 90.4 | 67.5     \\ \hline

\end{tabular}
\label{additional}
\end{table}


\section{DETAILS OF COMPARED METHODS}
\label{compared}
The details of compared methods are described following
\begin{itemize}[leftmargin=*]
    \item \textbf{SBA-Samp} \cite{zhang2021backdoor}: This method introduces a static subgraph as a trigger into the training graph for each poisoned node. The subgraph's connections are formed based on the Erdos-Renyi (ER) model, while its node features are randomly selected from those in the training graph.
    \item \textbf{SBA-Gene}: An adaptation of SBA-Samp, SBA-Gen differentiates itself by employing synthetically generated features for the trigger nodes. These features are drawn from a Gaussian distribution, the parameters of which—mean and variance—are derived from the attributes of actual nodes.
    \item \textbf{GTA} \cite{xi2021graph}: GTA utilizes a trigger generator that crafts subgraphs as triggers tailored to individual samples. The optimization of the trigger generator focuses exclusively on the backdoor attack loss, disregarding any constraints related to trigger detectability.
    \item \textbf{UGBA} \cite{Dai_2023}: UGBA select representative and diverse nodes as poisoned nodes to fully utilize the attack budget. An adaptive trigger generator is optimized with an constraint loss so that the generated triggers are ensured to be similar to the target nodes.
    
\end{itemize}

\section{IMPLEMENTATION DETAILS}
\label{impledetail}
A 2-layer GCN is utilized as the surrogate model, another 2-layer GCN is used for $f_d$, while a 2-layer MLP serves as the in-distribution trigger generator. We set all hidden dimensions to 256. The number of inner iteration steps, N and K, are consistently set to 1 and 20 across all experiments. 
The hyperparameters $\alpha$ and $\beta$ are selected based on the grid search on the validation set.
For the \textbf{OD} defense, the pruning threshold is set to exclude roughly $3\%$ of the samples with the highest reconstruction loss.

\section{TIME COMPLEXITY ANALYSIS}
\label{time}
During the bi-level optimization phase, the computation cost of each outter iteration consist of updating of surrogate GCN model and OOD detector in inner iterations and training adaptive trigger generator.
Let $h$ denote the embedding dimension. The cost for updating the surrogate model is approximately $O(Nhd|\mathcal{V}|)$, where $d$ is the average degree of nodes and $N$ is the number of inner iterations for the surrogate model, which is generally small. 
The cost for updating the OOD detector is approximately $O(Khd(|\mathcal{V}_s|+|\mathcal{T}_P|))$, where $K$ is the number of inner iterations for the OOD detector. 
For trigger generator, the cost for optimizing $\mathcal{L}_T$ is $O(hd|\mathcal{V}_U|)$, for optimizing $\mathcal{L}_D$ is $O(hd(|\mathcal{V}_s|+|\mathcal{T}_P|))$, and for optimizing $\mathcal{L}_E$ is $O(hd|\mathcal{V}|+|\mathcal{V}_U|^2h+|\mathcal{V}_U||\mathcal{V}_t||\mathcal{V}_L|h)$, where $|\mathcal{V}_L|$ and $|\mathcal{V}_t|$ are generally small compared to $|\mathcal{V}|$.
In our empirical experiments conducted on large-scale datasets, such as Flickr and OGB-arxiv, which comprise 899,756 and 169,343 nodes respectively, we streamlined the training process by selecting a subset of $\mathcal{V}_U$ and setting $|\mathcal{V}_U|=4096$ for each epoch. Despite this simplification, DPGBA still achieve a high attack success rate, as evidenced in Tab. \ref{GCN}~--~\ref{GAT}.
In Table \ref{trainingtime}, we report the overall training time and corresponding ASR of our DPGBA compared to GTA and UGBA on the OGB-arxiv dataset. All models were trained on a Nvidia A6000 GPU with 48GB of memory. 
The results indicate that DPGBA requires only approximately 20 seconds more training time compared to the baselines on the OGB-arxiv dataset.
Given that our DPGBA achieves an ASR of over 90\%, while the baseline methods achieve nearly 0\% with OD defense adopted, this additional time is justified. 
This demonstrates that DPGBA effectively generates triggers that the victim model quickly memorizes, highlighting its potential for conducting scalable targeted attacks.

\begin{table}[t]
\centering
\caption{Training Time}
\vskip -1em
\begin{tabular}{cccc}
\hline
\textbf{Metrics} & \textbf{GTA} & \textbf{UGBA} & \textbf{DPGBA} \\ \hline
\textbf{ASR (None)} & 92.0 & 94.3 & 92.7 \\ 
\textbf{ASR (OD)} & 0.00 & 0.00 & 92.0 \\ 
\textbf{Time} & 37.7s & 41.8s & 60.6s \\ \hline
\end{tabular}
\label{trainingtime}
\end{table}

\end{document}